\documentclass[letterpaper, 10 pt, conference]{ieeeconf}  

\IEEEoverridecommandlockouts                              

\usepackage{graphicx} 
\usepackage[FIGBOTCAP, TABBOTCAP]{subfigure}
\usepackage{amsmath} 
\usepackage{amssymb}  
\usepackage{xcolor}
\usepackage{tabularx}
\usepackage{multirow}
\usepackage{mathtools}
\usepackage{cuted}

\graphicspath{ {./figures/} }

\title{\LARGE \bf
  Solving the swing-up and balance task for the Acrobot and Pendubot with SAC
}

\author{Chi Zhang$^{1,2}$, Akhil Sathuluri$^{1}$, Markus Zimmermann$^{1}$
    \thanks{$^{1}$Technical University of Munich, Laboratory for Product Development and Lightweight Design, Munich, Germany  {\tt\small chi97.zhang@mytum.de , akhil.sathuluri@tum.de}}
    \thanks{$^{2}$Robotics Innovation Center, DFKI GmbH, Bremen, Germany}
    \thanks{This work was supported by M-RoCK (FKZ 01IW21002) project funded by the German Aerospace Center (DLR) with federal funds from the Federal Ministry of Education and Research (BMBF).}
}

\begin{document}

\maketitle
\thispagestyle{empty}
\pagestyle{empty}

\begin{abstract}
  We present a solution of the swing-up and balance task for the pendubot and
  acrobot for the participation in the AI Olympics competition at IJCAI 2023.
  Our solution is based on the Soft Actor Crtic (SAC) reinforcement learning
  (RL) algorithm for training a policy for the swing-up and entering the region
  of attraction of a linear quadratic regulator (LQR) controller for
  stabilizing the double pendulum at the top position. Our controller achieves
  competitive scores in performance and robustness for both, pendubot and
  acrobot, problem scenarios.
\end{abstract}

\section{Introduction}
\label{sec_introduction}

Robotic control is a complex problem and is approached in many different ways,
such as classical motion planning, optimal control methods and reinforcement
learning methods. The AI Olympics competition at IJCAI 2023 is an attempt to
compare control methods in a simple but challenging setting. The competition
involves the application of the control methods not only in a simulation setting
but also on a real combined pendubot and acrobot hardware system.

Extensive research has been conducted on complex robotic motion control, with
learning-based methods gaining widespread adoption in recent times. A notable
example is the classical control module within the Gymnasium
environment~\cite{brockman2016openai}, a well-known reinforcement learning
framework developed by OpenAI.  This module serves as an ideal platform for
exploring and evaluating new motion control algorithms. Another noteworthy
achievement in the field is the DeepMimic project~\cite{2018-TOG-deepMimic}.
Through imitation learning within a simulated environment, this project has
enabled robots to acquire intricate human movements, including backflips and
martial arts maneuvers. Additionally, a research team from ETH Zurich has made
significant progress in addressing the sim-to-real interface
challenge~\cite{hwangbo2019learning}. Their methodology involves training a
control model for a mini-cheetah-like robot within a simulation environment. By
utilizing a neural network and leveraging data collected from the real robot,
they approximated the dynamics model of the physical robot. This approach has
facilitated accurate implementation of the control policy derived from the
virtual environment onto the real robot. These exemplary works demonstrate the
promising applications of learning-based approaches in various aspects of robot
control.


The AI Olympics challenge focuses on two underactuated variations of the double
pendulum model, namely the pendubot and acrobot. The objective revolves around
accomplishing two primary tasks: swinging the double pendulum from its lowest
position to its highest point, and maintaining stability at the highest point.
To achieve the swing-up task, we employ the classical model-free reinforcement
learning algorithm known as Soft Actor-Critic (SAC)~\cite{haarnoja2018soft} to
train a policy which is able to reach the region of attraction (RoA) of a continuous time linear quadratic regulator (LQR) controller~\cite{Lehtomaki1981,underactuated}. As
soon as the system enters the RoA, we transition to the LQR controller to
stabilize the entire system.

This paper is structured as follows: In the next section, we briefly summarize
the SAC algorithm. In Section~\ref{sec:controller}, we explain the reward
function that we used for the training, how we conducted the training and how we
composed the controller. We present the results in Section~\ref{sec:results} and
conclude in Section~\ref{sec:conclusion}.

\section{Soft-Actor-Crtic}
Soft Actor Critic (SAC) ~\cite{haarnoja2018soft} is a popular algorithm
used in the field of reinforcement learning. It is originally
designed for continuous action spaces, where the agent has
an infinite choice of actions to take. In our problem scenario,
the actuators of the double pendulum can be set to any
value within the torque limit range. The position and velocity
measurements obtained from the motors are also represented
as continuous real numbers. Therefore, we choose the SAC algorithm to train the agent.

SAC optimizes a policy by maximizing the expected cumulative reward obtained by
the agent over time. This is achieved through an actor and critic
structure~\cite{konda1999actor}.

The actor is responsible for selecting actions based on the current policy in
response to the observed state of the environment. It is typically represented
by a shallow neural network that approximates the mapping between the input
state and the output probability distribution over actions. SAC incorporates a
stochastic policy in its actor part, which encourages exploration and helps the
agent improve policies.

The critic, on the other hand, evaluates the value of state-action pairs. It
estimates the expected cumulative reward that the agent can obtain by following
a certain policy. Typically, the critic is also represented by a neural
network that takes state-action pairs as inputs and outputs estimated value.

In addition to the actor and critic, a central feature of SAC is entropy
regularization\cite{SpinningUp2018}. The policy is trained to maximize a trade-off between expected return and entropy, which is a measure of randomness in the action selection. If
\(x\) is a random variable with a probability density function \(P\), the
entropy \(H\) of \(x\) is defined as:

\[
 H(P) = \displaystyle \mathop{\mathbb{E}}_{x \sim P}[-\log P(x)]
\]

By maximizing entropy, SAC encourages exploration and accelerates learning. It
also prevents the policy from prematurely converging to a suboptimal solution.
The trade-off between maximizing reward and maximizing entropy is controlled
through a parameter, \(\alpha\). This parameter serves to balance the importance
of exploration and exploitation within the optimization problem. The optimal policy
\(\pi^*\) can be defined as follows:

\[
 \pi^* = {arg}{\max_{\pi}}{\displaystyle
 \mathop{\mathbb{E}}_{\tau\sim\pi}}{\Bigg[{\sum_{t=0}^{\infty}}{\gamma^{t}}{\Big(R(s_t,a_t,s_{t+1})}+{\alpha}H(\pi(\cdot\mid{s_t}))\Big)\Bigg]}
\]

During training, SAC learns a policy $\pi_{\theta}$ and two Q-functions
$Q_{\phi_1} , Q_{\phi_2}$ concurrently. The loss functions for the two Q-networks are
$(i \in {1, 2})$:

\[
  L(\phi_i,D) = \displaystyle
  \mathop{\mathbb{E}}_{(s,a,r,s',d)\sim{D}}\bigg[\bigg(Q_{\phi_i}(s,a)-y(r,s',d)\bigg)^2\bigg],
\]

where the temporal difference target \(y\) is given by:
\begin{align*}
  y(r,s',d) &= r + \gamma(1-d) \times \nonumber \\
  & \bigg(\displaystyle
  \mathop{\min}_{j=1,2}Q_{\phi_{targ,j}}(s',\tilde{a}')-\alpha\log
  {\pi_\theta}(\tilde{a}'\mid{s}')\bigg), \\
  \tilde{a}'&\sim{\pi_\theta}(\cdot\mid{s'})
\end{align*}

In each state, the policy \(\pi_\theta\) should act to maximize the expected
future return \(Q\) while also considering the expected future entropy \(H\). In other
words, it should maximize \(V^\pi(s)\):
\begin{align*}
 V^\pi(s) &= {\displaystyle \mathop{\mathbb{E}}_{a\sim\pi}[Q^\pi(s,a)]} +
 \alpha{H(\pi(\cdot\mid{s}))} \\
 &= {\displaystyle \mathop{\mathbb{E}}_{a\sim\pi}[Q^\pi(s,a)]} -
 \alpha{\log {\pi(a\mid{s})}}
\end{align*}

By employing an effective gradient-based optimization technique, the parameters
of both the actor and critic neural networks undergo updates, subsequently
leading to the adaptation of the policies themselves.

In conclusion, SAC's combination of stochastic policies, exploration through
entropy regularization, value estimation, and gradient-based optimization make
it a well-suited algorithm for addressing the challenges posed by continuous
state and action spaces.

\section{Controller}
\label{sec:controller}
Pendubot and acrobot represent two variations of the double pendulum,
distinguished by which joint is actuated. In the case where the shoulder joint
is actuated, it is referred to as pendubot, while the term acrobot is used when
the elbow joint is actuated. To conduct our study, we used a custom
simulation environment building up on the simulation developed by the
competition organizers~\cite{2023_ram_wiebe_double_pendulum}. The environment is constructed based on the dynamics provided in the double pendulum repository for the competition, and it incorporates standard OpenAI Gym environment features.
We used the vanilla SAC algorithm implemented in Stable Baseline3~\cite{stable-baselines3}.

The simulation environment encompasses four controlled variables: the angular position ($p_1$) and angular velocity ($v_1$) of the shoulder joint, as well as the
angular position ($p_2$) and angular velocity ($v_2$) of the elbow joint. The
whole state is $x = [p_q, p_2, v_1, v_2]^T$. The
controller generates output $u$ which corresponds to the torque applied on the
shoulder \(\tau_1\) for the pendubot
and to the torque applied on the elbow \(\tau_2\) for the acrobot. In order to
facilitate effective training of the agent, we found it beneficial to use
normalized state and action representations. This mapping between real-world
physical quantities \((x,u)\) and the agent's state and action space \((s,a)\)
are

\begin{align}
  a &= \tau_{max} u\\
  s_i &= \frac{(p_i \mod 2\pi) - \pi}{\pi} \quad , i \in \{1, 2\}\\
  s_i &= \frac{\min (\max (x_i , -v_{max}), v_{max} )}{v_{max}} \quad , i \in \{3, 4\}
\end{align}

We used $\tau_{max} = 5.0\,\text{Nm}$ and $v_{max} = 20.0 \,
\text{rad}/\text{s}$.

In theory, the reward function is designed to guide the agent's behavior towards
achieving stability around the system's goal point. However, practical
implementation has revealed challenges such as potential entrapment in local
minima or difficulty in maintaining stability at the highest point for extended
periods. To address these issues, we employ two main approaches. For the
stabilization problem, we introduce the concept of a combined controller. During
the swing-up process, the RL-trained agent assumes control, leveraging its
learned policies. However, as the system approaches proximity to the maximum
point, a smooth transition occurs, allowing a continuous time LQR controller to provide the
final stabilization necessary for maintaining stability at the highest point.

To steer the agent away from hazardous local minima, we devised a three-stage
reward function. The full equation for this reward function is

\begin{align}
  r(x,u) = &-(x - x_g)^T Q_{train} (x - x_g) - u^T R_{train}u \nonumber\\
           & +
            \begin{dcases*}
              r_{line} & \text{if} $h(p_1, p_2) \geq h_{line}$\, ,\\
              0 & \text{else}
            \end{dcases*}\nonumber\\
           & +
            \begin{dcases*}
              r_{LQR} & \text{if} $(x - x_g)^T S_{LQR} (x - x_g) \geq \rho $\, ,\\
              0 & \text{else}
            \end{dcases*}\nonumber\\
           & -
            \begin{dcases*}
              r_{vel} & \text{if} $|v_1| \geq v_{thresh}$\, ,\\
              0 & \text{else}
            \end{dcases*}\nonumber\\
           & -
            \begin{dcases*}
              r_{vel} & \text{if} $|v_2| \geq v_{thresh}$\, ,\\
              0 & \text{else}
            \end{dcases*}
\end{align}

In the initial stage, a quadratic reward function is employed to encourage
smooth swinging of the entire system within a relatively small number of
training sessions. The matrix  \(Q_{train} = diag(Q_1, Q_2, Q_3, Q_4)\) is a
diagonal matrix, while \(R_{train}\) is a scalar. This is due to the nature
of underactuated control in the double pendulum system, where only a single
control input is available.

As the end effector reaches a threshold line \(h_{line} = 0.8(l_1+l_2)\), we
introduce a second level of reward \(r_{line}\). The end effector height is
given by
\begin{align}
    h(p_1, p_2) = -l_1\cos(p_1) - l_2 \cos(p_1 + p_2).
\end{align}
with the link lengths $l_1$ and $l_2$.
This reward provides the agent with a fixed value
but is carefully designed to prevent the system from spinning rapidly in either
clockwise or counterclockwise directions. To discourage the agent from
exploiting rewards by spinning at excessive speeds, a significant penalty
\(-r_{vel}\) is implemented for any speed exceeding $v_{thresh}=8\,
\text{rad}/\text{s}$ in absolute value.
This penalty effectively compels the agent to approach the maximum point while
adhering to the predefined speed interval. The speed penalty was only needed for
the acrobot.

The third level of reward $r_{LQR}$ aims to provide a substantial reward to the
agent when it remains within the Region of Attraction (ROA) of the LQR
controller. By this we want to achieve that the policy learns to enter the LQR
controller RoA so that there can be a smooth transition between both
controllers. For details on the LQR controller and its region of attraction, we
refer to these lecture notes~\cite{underactuated}. The parameters, we used in
the cost matrices of the LQR controller are listed in
Table~\ref{tab:parameters}. We computed the RoA similar to~\cite{maywald2022}
but with a sums of squares method~\cite{tedrake2010}.  Once the RoA is computed,
it can be checked whether a state $x$ belongs to the estimated RoA of the LQR
controller by calculating the cost-to-go of the LQR controller with the matrix
$S_{LQR}$ and comparing it with the scalar $\rho$.

The parameters we used for the reward function and the LQR controller are listed
in Table~\ref{tab:parameters}.

\begin{table}
  \centering
  \begin{tabular}{p{1.5cm} |p{1.5cm} p{1.8cm} p{1.5cm}}
  Robot & Quadratic Reward  & Constant Reward & LQR\\
  \hline
  \multirow{5}{*}{Pendubot} & \(Q_1\) = 8.0  &  & \(Q_1\) = 1.92\\
  & \(Q_2\) = 5.0  & \(r_{line}=500\) & \(Q_2\) = 1.92\\
  & \(Q_3\) = 0.1  & \(r_{vel}=0.0\) & \(Q_3\) = 0.3\\
  & \(Q_4\) = 0.1  & \(r_{LQR}=1e4\)& \(Q_4\) = 0.3\\
  & \(R\) = 1e-4  & & \(R\) = 0.82\\
  \hline
  \multirow{5}{*}{Acrobot} & \(Q_1\) = 10.0  &  & \(Q_1\) = 0.97\\
  & \(Q_2\) = 10.0  & \(r_{line}=500\) & \(Q_2\) = 0.93\\
  & \(Q_3\) = 0.2  & \(r_{vel}=1e4\) & \(Q_3\) = 0.39\\
  & \(Q_4\) = 0.2  & \(r_{LQR}=1e4\) & \(Q_4\) = 0.26\\
  & \(R\) = 1e-4  &  & \(R\) = 0.11\\
  \end{tabular}
 \caption{Hyper parameters used for the SAC training and the LQR controller.}
 \label{tab:parameters}
\end{table}

During the training phase of the SAC controller for both the acrobot and
pendubot, we placed our focus on tuning several crucial hyperparameters. These
hyperparameters included the learning rate, control frequency, episode length,
and learning time steps. We carefully set the learning rate to 0.01 to
facilitate effective learning and adaptation. Additionally, the control
frequency of the simulation was set to 100Hz, ensuring frequent updates and responsiveness in the control process. For each training episode, we chose an episode length of 1000
(Acrobot) and 500 (Pendubot) corresponding to $10\,\text{s}$ and $5\,\text{s}$
long episodes to provide sufficient exploration and learning opportunities. To
maximize the training potential, we conducted a total of $2e7$ learning time
steps, allowing the agent to gather extensive experience and refine its
performance.

During the execution, at every control step, it is checked whether the current
state belongs to the estimated RoA of the LQR controller. If it does the LQR
controller command is used else the SAC policy command is used.

We encountered heavy oscillation in the simulation when the
LQR controller took over from the SAC controller. This oscillation was primarily
caused by the low control frequency. To address this issue and maintain
stability, we adjusted the control frequency to 500Hz, enabling smoother control
and reducing the oscillation experienced by the LQR controller.

\section{Results}
\label{sec:results}
\begin{figure}[h]
\centering
 \includegraphics[scale = 0.2]{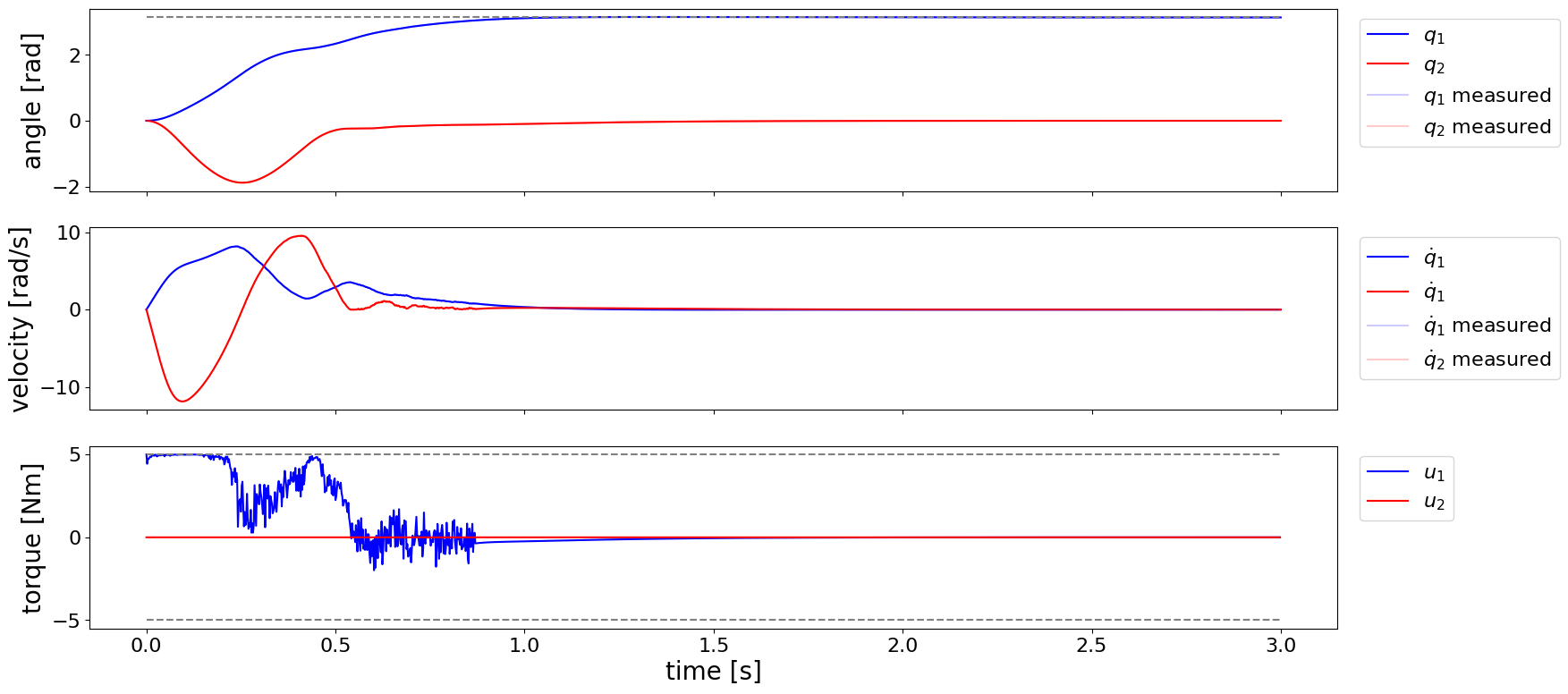}
 \caption{Swing-up trajectory of the pendubot.}
 \label{fig:timeseries_pendubot}
\end{figure}

Our carefully designed pendubot controller delivered commendable outcomes in
both the swing-up and stabilization tasks. With a swing-up time metric of 0.65
seconds, our pendubot controller achieved a performance level that matches the
top result on the leaderboard by the time-varying LQR controller. The swing-up
trajectory is visualized in Fig.  \ref{fig:timeseries_pendubot}. Most other
performance metrics (Table \ref{tab:performance}), such as Energy, Integrated
Torque, and Velocity Cost, demonstrated impressive performance, with the
exception of Torque smoothness, where the controller showed some room for
improvement.

\begin{table}
  \centering
 \begin{tabular}{lcc}
  Criteria& Pendubot  & Acrobot \\
 \hline
 Swingup Success& success  & success\\
 Swingup time [s]& 0.65   & 2.06\\
 Energy [J]& 9.4  & 29.24 \\
 Max. Torque [Nm]& 5.0   & 5.0 \\
 Integrated Torque [Nm]& 2.21  & 4.57 \\
 Torque Cost [N²m²] & 8.58  & 12.32 \\
 Torque Smoothness [Nm]& 0.172 & 0.954 \\
 Velocity Cost [m²/s²]& 44.98 & 193.78  \\
 RealAI Score & 0.801 & 0.722  \\
 \end{tabular}
 \caption{Performance scores of our controller for pendubot and acrobot.}
 \label{tab:performance}
\end{table}

\begin{figure}[h]
\centering
\subfigure[Pendubot]{\includegraphics[width=0.45\linewidth]{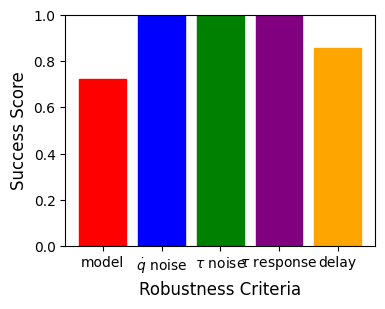}}
\hfill
\subfigure[Acrobot]{\includegraphics[width=0.45\linewidth]{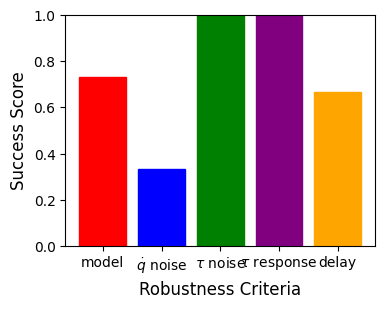}}
\caption{Robustness results of our controller for pendubot and acrobot.}
\label{fig:robustness}
\end{figure}

In terms of robustness, the controller surpassed our expectations with a score
of 0.916 (Fig.  \ref{fig:robustness}(a)). It demonstrated near-perfect
resilience to measurement noise, torque noise, and torque response.
Additionally, the controller outperformed the majority of other controllers when
faced with challenges such as time delays and model inaccuracies.
The combined score of performance and robustness for the pendubot is 0.86.

\begin{figure}[h!]
\centering
 \includegraphics[scale = 0.2]{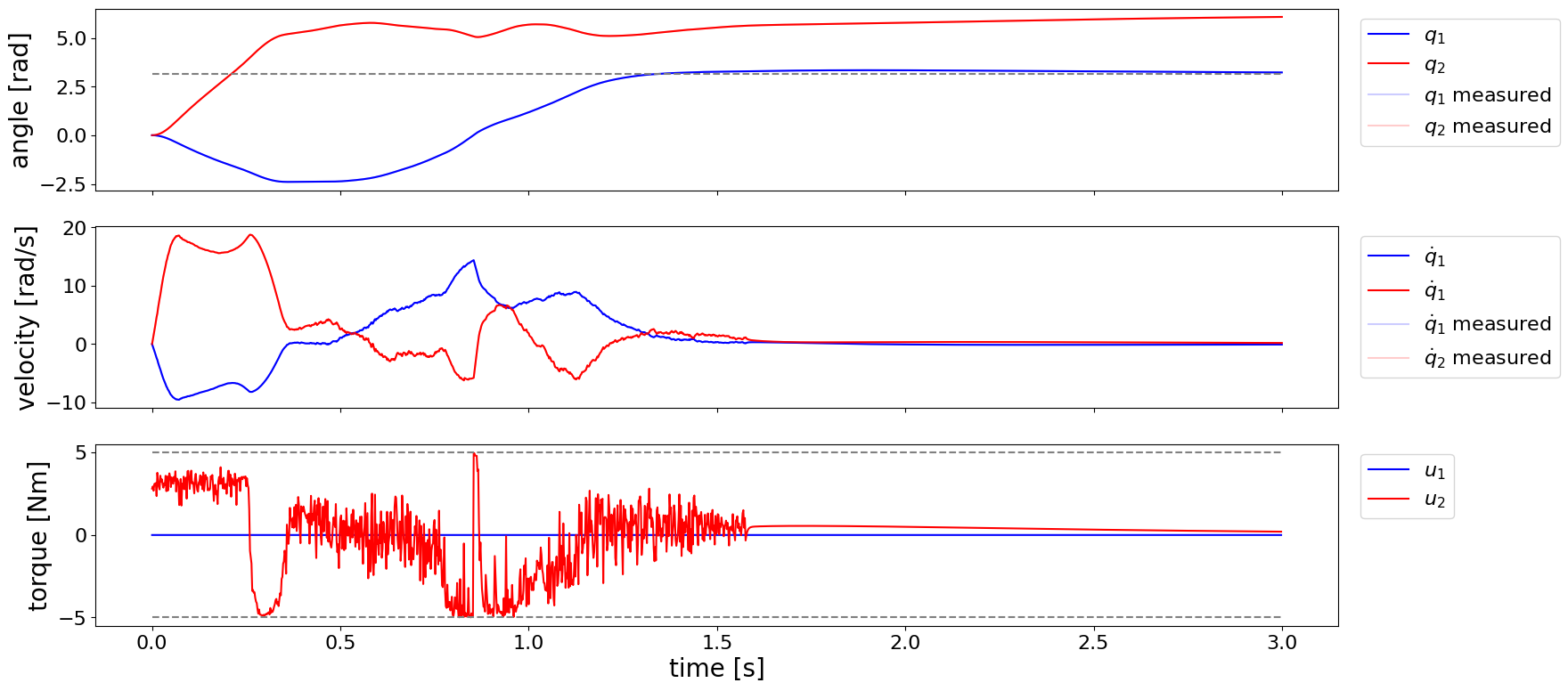}
 \caption{Swing-up trajectory of the Acrobot.}
 \label{fig:timeseries_acrobot}
\end{figure}

The performance evaluation of our trained model on the acrobot task reveals its
success in both swing-up and stabilization tasks (trajectory in Fig.
\ref{fig:timeseries_acrobot}, performance scores in Table
\ref{tab:performance}). Particularly noteworthy is its
swing-up time of only 2.06 seconds, giving it a clear advantage in terms of
speed. Although our model delivers fair results in terms of maximum torque,
integrated torque, and torque cost, we have made some compromises in terms of
torque smoothness and energy efficiency.


When it comes to robustness, our trained model proves resilient against torque
noise and demonstrates commendable performance in torque response (Fig.
\ref{fig:robustness}(b)). It also holds
up reasonably well against model inaccuracies and time delays. However, it
exhibits high sensitivity when encountering measurement noise.
The overall robustness score is 0.747 and the average with the performance score
is 0.73.

\section{Conclusion}
\label{sec:conclusion}

The combination of an RL-based controller and a continuous time LQR controller
proved to be successful in controlling both the acrobot and pendubot systems for
swing-up and stabilization tasks. Notably, both variations exhibited commendable
swing-up time metrics and demonstrated robustness. However, there were two main
challenges encountered during the experiment.

The first challenge involved establishing a smooth switching condition between
the RL-based controller and the LQR controller. This necessitated optimizing the
region of attraction specific to the LQR controller, ensuring a seamless
transition between the two control approaches. The second challenge revolved
around training a stable agent using a model-free approach, specifically SAC,
to ensure that the entire system approached the target
point at a sufficiently low speed to enter the region of attraction where the
LQR controller could take over.

To overcome these challenges, we developed a three-stage reward function to
interface with the agent. However, the process of fine-tuning the
hyperparameters for this reward function proved to be quite time-consuming.

Throughout our experiments, we have come very close to relying solely on the
RL-based controller for both swing-up and stabilization tasks. In the future,
our plans involve enhancing the existing reward function, which will increase
the likelihood of training an agent capable of achieving self-stabilization
around a desired position. Additionally, we will focus our efforts on further
approximating the training environment to closely resemble real-world physics.
This will facilitate a smoother transition of the controller from the simulated
environment to the actual physical system.






\bibliographystyle{IEEEtran}
\bibliography{references}

\begin{thebibliography}{10}
\providecommand{\url}[1]{#1}
\csname url@samestyle\endcsname
\providecommand{\newblock}{\relax}
\providecommand{\bibinfo}[2]{#2}
\providecommand{\BIBentrySTDinterwordspacing}{\spaceskip=0pt\relax}
\providecommand{\BIBentryALTinterwordstretchfactor}{4}
\providecommand{\BIBentryALTinterwordspacing}{\spaceskip=\fontdimen2\font plus
\BIBentryALTinterwordstretchfactor\fontdimen3\font minus
  \fontdimen4\font\relax}
\providecommand{\BIBforeignlanguage}[2]{{%
\expandafter\ifx\csname l@#1\endcsname\relax
\typeout{** WARNING: IEEEtran.bst: No hyphenation pattern has been}%
\typeout{** loaded for the language `#1'. Using the pattern for}%
\typeout{** the default language instead.}%
\else
\language=\csname l@#1\endcsname
\fi
#2}}
\providecommand{\BIBdecl}{\relax}
\BIBdecl

\bibitem{brockman2016openai}
G.~Brockman, V.~Cheung, L.~Pettersson, J.~Schneider, J.~Schulman, J.~Tang, and
  W.~Zaremba, ``Openai gym,'' \emph{arXiv preprint arXiv:1606.01540}, 2016.

\bibitem{2018-TOG-deepMimic}
\BIBentryALTinterwordspacing
X.~B. Peng, P.~Abbeel, S.~Levine, and M.~van~de Panne, ``Deepmimic:
  Example-guided deep reinforcement learning of physics-based character
  skills,'' \emph{ACM Trans. Graph.}, vol.~37, no.~4, pp. 143:1--143:14, Jul.
  2018. [Online]. Available: \url{http://doi.acm.org/10.1145/3197517.3201311}
\BIBentrySTDinterwordspacing

\bibitem{hwangbo2019learning}
J.~Hwangbo, J.~Lee, A.~Dosovitskiy, D.~Bellicoso, V.~Tsounis, V.~Koltun, and
  M.~Hutter, ``Learning agile and dynamic motor skills for legged robots,''
  \emph{Science Robotics}, vol.~4, no.~26, p. eaau5872, 2019.

\bibitem{haarnoja2018soft}
T.~Haarnoja, A.~Zhou, P.~Abbeel, and S.~Levine, ``Soft actor-critic: Off-policy
  maximum entropy deep reinforcement learning with a stochastic actor,'' in
  \emph{International conference on machine learning}.\hskip 1em plus 0.5em
  minus 0.4em\relax PMLR, 2018, pp. 1861--1870.

\bibitem{Lehtomaki1981}
N.~Lehtomaki, N.~Sandell, and M.~Athans, ``Robustness results in
  linear-quadratic gaussian based multivariable control designs,'' \emph{IEEE
  Transactions on Automatic Control}, vol.~26, no.~1, pp. 75--93, 1981.

\bibitem{konda1999actor}
V.~Konda and J.~Tsitsiklis, ``Actor-critic algorithms,'' \emph{Advances in
  neural information processing systems}, vol.~12, 1999.

\bibitem{SpinningUp2018}
J.~Achiam, ``{Spinning Up in Deep Reinforcement Learning},'' 2018.

\bibitem{2023_ram_wiebe_double_pendulum}
F.~Wiebe, S.~Kumar, L.~Shala, S.~Vyas, M.~Javadi, and F.~Kirchner, ``An open
  source dual purpose acrobot and pendubot platform for benchmarking control
  algorithms for underactuated robotics,'' \emph{IEEE Robotics and Automation
  Magazine}, 2023, under review.

\bibitem{stable-baselines3}
\BIBentryALTinterwordspacing
A.~Raffin, A.~Hill, A.~Gleave, A.~Kanervisto, M.~Ernestus, and N.~Dormann,
  ``Stable-baselines3: Reliable reinforcement learning implementations,''
  \emph{Journal of Machine Learning Research}, vol.~22, no. 268, pp. 1--8,
  2021. [Online]. Available: \url{http://jmlr.org/papers/v22/20-1364.html}
\BIBentrySTDinterwordspacing

\bibitem{underactuated}
\BIBentryALTinterwordspacing
R.~Tedrake, \emph{Underactuated Robotics}, 2022. [Online]. Available:
  \url{http://underactuated.mit.edu}
\BIBentrySTDinterwordspacing

\bibitem{maywald2022}
L.~Maywald, F.~Wiebe, S.~Kumar, M.~Javadi, and F.~Kirchner, ``Co-optimization
  of acrobot design and controller for increased certifiable stability,'' in
  \emph{IEEE/RSJ International Conference on Intelligent Robots and Systems
  (IROS-2022)}, 10 2022.

\bibitem{tedrake2010}
R.~Tedrake, I.~R. Manchester, M.~Tobenkin, and J.~W. Roberts, ``{{LQR-trees}}:
  {{Feedback Motion Planning}} via {{Sums-of-Squares Verification}},''
  \emph{The International Journal of Robotics Research}, vol.~29, no.~8, pp.
  1038--1052, Jul. 2010.

\end{thebibliography}


\end{document}